\documentclass[twoside,leqno,twocolumn]{article}

\usepackage[letterpaper]{geometry}

\usepackage{ltexpprt}
\usepackage{hyperref}

\usepackage{amsmath,amssymb,amsfonts}

\usepackage{url}            
\usepackage{amsfonts}       
\usepackage{microtype}
\usepackage{graphicx}
\graphicspath{{figures/}}
\usepackage{subcaption}
\usepackage{colortbl}
\usepackage{mathtools}
\usepackage{multirow}
\usepackage{tabularx}
\usepackage{float}
\usepackage{algorithmic}
\usepackage{algorithm}
\usepackage[dvipsnames]{xcolor}

\usepackage[inline]{enumitem}
\usepackage[english]{babel}
\usepackage{wrapfig}
\usepackage{color,soul}
\usepackage{changepage}
\usepackage{amssymb}
\DeclareMathOperator*{\argmin}{argmin} 
\usepackage{bbm}

\begin{document}

\newcommand\relatedversion{}
\renewcommand\relatedversion{\thanks{The full version of the paper can be accessed at \protect\url{https://arxiv.org/abs/1902.09310}}} 

\title{\Large Saliency-Augmented Memory Completion for Continual Learning}
\author{Guangji Bai\thanks{Emory University, Atlanta, GA. \{guangji.bai, chen.ling, yuyang.gao, liang.zhao\}@emory.edu.}
\and Chen Ling\footnotemark[1]
\and Yuyang Gao\footnotemark[1]
\and Liang Zhao\footnotemark[1]}

\date{}

\maketitle


\fancyfoot[R]{\scriptsize{Copyright \textcopyright\ 2023 by SIAM\\
Unauthorized reproduction of this article is prohibited}}





\begin{abstract} \small\baselineskip=9pt 
Continual Learning (CL) is considered a key step toward next-generation Artificial Intelligence. Among various methods, replay-based approaches that maintain and replay a small episodic memory of previous samples are one of the most successful strategies against catastrophic forgetting. However, since forgetting is inevitable given bounded memory and unbounded tasks, ‘how to forget’ is a problem continual learning must address. Therefore, beyond simply avoiding (catastrophic) forgetting, an under-explored issue is how to reasonably forget while ensuring the merits of human memory, including 1) storage efficiency, 2) generalizability, and 3) some interpretability. To achieve these simultaneously, our paper proposes a new saliency-augmented memory completion framework for continual learning, inspired by recent discoveries in memory completion/separation in cognitive neuroscience. Specifically, we innovatively propose to store the part of the image most important to the tasks in episodic memory by saliency map extraction and memory encoding. When learning new tasks, previous data from memory are inpainted by an adaptive data generation module, which is inspired by how humans ``complete'' episodic memory. The module's parameters are shared cross all tasks and it can be jointly trained with a continual learning classifier as bilevel optimization. Extensive experiments on several continual learning and image classification benchmarks demonstrate the proposed method's effectiveness and efficiency.
\end{abstract}

\section{Introduction}
\label{sec:intro}

An important step toward next-generation Artificial Intelligence (AI) is a promising new domain known as continual learning (CL), where neural networks learn continuously over a sequence of tasks, similar to the way humans learn~\cite{parisi2019continual}. Compared with traditional supervised learning, continual learning is still in its very primitive stage. Currently, the primary goal is essentially to avoid \textit{Catastrophic Forgetting}~\cite{mccloskey1989catastrophic} of previously learned tasks when an agent is learning new tasks. Continual learning aims to mitigate forgetting while updating the model over a stream of tasks. 



To overcome this issue, researchers have proposed a number of different strategies. Among various approaches, the \textit{replay-based methods} are arguably more effective in terms of performance and bio-inspiration~\cite{robins1995catastrophic} as a way to alleviate the catastrophic forgetting challenge and are thus becoming the preferred approach for continual learning models~\cite{rebuffi2017icarl,aljundi2019gradient}. However, the performance of these methods highly depends on the size of the episodic memory. Recent work~\cite{knoblauch2020optimal} proved that to achieve optimal performance in continual learning, one has to store all previous examples in the memory, which is almost impossible in practice and counters the way how human brain works. Unlike avoiding (catastrophic) forgetting, the attempts on `how to reasonably forget' is still a highly open question, leading to significant challenges in continual learning, including \textbf{1) Memory inefficiency.} The performance of replay-based models depends heavily on the size of the available memory in the replay buffer, which is used to retain as many previous samples as possible. While existing works typically store the entire sample in memory, we humans seldom memorize every detail of our experiences. Thus, compared to biological neural networks, some mechanisms must still be missing in current models; \textbf{2) Insufficient generalization power.} The primary focus of existing works is to avoid catastrophic forgetting by memorizing all the details
without taking into account their usefulness for learning tasks. They typically rely on episodic memory for individual tasks without sufficient chaining to make the knowledge they learn truly generalizable to all potential (historical and future unseen) tasks. In contrast, human beings significantly improve generalizability during continual learning; \textbf{3) Obscurity of the memory and its importance to learning tasks.} Human being usually has a concise and clear clue on how the relevant memory is useful for the learning tasks. Such a clue could even be helpful for telling if the human has the necessary memory to be capable of a specific task at all. However, the majority of existing works in CL pay little attention to pursuing such transparency of continual learning models. 
For example, most works directly feed the images into their model for training without any explanation generated, which may prevent users from understanding which semantic features in the image are the most decisive ones and how the model is reasoning to make the final prediction. 
In addition, without an explanation generation mechanism, it is much more challenging for model diagnosing or debugging, let alone understanding how knowledge is stored and refined through the continual learning process.

To jointly address these challenges, this paper proposes \textbf{\underline{S}}aliency-\textbf{\underline{A}}ugmented \textbf{\underline{M}}emory \textbf{\underline{C}}ompletion (\textbf{SAMC}) framework for continual learning, which is inspired by the \emph{memory pattern completion} theory in cognitive neuroscience. The memory pattern completion process guides the abstraction of learning episodes (tasks) into semantic knowledge as well as the reverse process in recovering episodes from the memorized abstracts~\cite{hunsaker2013operation}. Specifically, in this paper, instead of memorizing all historical training samples, we memorize their interpretable abstraction in terms of the saliency maps that most determine the prediction outputs for each learning task. 
\textbf{Our contribution includes,} 1). We develop a novel neural-inspired continual learning framework to handle the catastrophic forgetting. 2). We propose two techniques based on saliency map and image inpainting methods for efficient memory storage and recovery. 3). We design a bilevel optimization algorithm with theoretical guarantee to train our entire framework in an end-to-end manner. 4). We demonstrate our model's efficacy and superiority with extensive experiments.

\section{Related Work}

\textbf{Continual Learning (CL).} Catastrophic forgetting is a long-standing problem~\cite{robins1995catastrophic} in continual learning which has been recently tackled in a variety of visual tasks such as image classification~\cite{kirkpatrick2017overcoming,rebuffi2017icarl}, object detection~\cite{zhou2020lifelong}, etc.

Existing techniques in CL can be divided into three main categories~\cite{parisi2019continual}: 1) \emph{regularization-based approaches}, 2) \emph{dynamic architectures} and 3) \emph{replay-based approaches}. Regularization-based approaches alleviates catastrophic forgetting by either adding a regularization term to the objective function~\cite{kirkpatrick2017overcoming} or knowledge distillation over previous tasks~\cite{li2017learning}. Dynamic architecture approaches adaptively accommodate the network architecture (e.g., adding more neurons or layers) in response to new information during training. Dynamic architectures can be explicit, if new network branches are grown, or implicit, if some network parameters are only available for certain tasks. Replay-based approaches alleviate the forgetting of deep neural networks by replaying stored samples from the previous history when learning new ones, and has been shown to be the most effective method for mitigating catastrophic forgetting. 

Replay-based methods mainly include three directions: namely rehearsal methods, constrained optimization and pseudo rehearsal~\cite{delange2021continual}. \textit{Rehearsal methods} directly retrieve previous samples from a limited size memory together with new samples for training~\cite{chaudhry2019tiny,rolnick2018experience}. While simple in nature, this approach is prone to overfitting the old samples from the memory. As an alternative, \textit{constrained optimization} methods formulate backward/forward transfer as constraints in the objective function. GEM~\cite{lopez2017gradient} constrains new task updates to not interfere with previous tasks by projecting the estimated gradient on the feasible region outlined by previous task gradients through first order Taylor series approximation. A-GEM~\cite{chaudhry2018efficient} further extended GEM and made the constraint computationally more efficient. EPR~\cite{saha2021saliency} uses zero-padding for memory efficiency while lacks theoretical guarantee on its performance. Finally, \textit{pseudo-rehearsal} methods typically utilize generative model such as GAN~\cite{goodfellow2020generative} or VAE~\cite{pu2016variational} to generate previous samples from random inputs and have shown the ability to generate high-quality images recently~\cite{robins1995catastrophic}. Readers may refer to~\cite{parisi2019continual,delange2021continual} for a more comprehensive survey on continual learning.


\noindent\textbf{Saliency Detection.} Saliency detection is to identify the most informative part of input features. It has been applied to various domains including CV~\cite{goferman2011context}, NLP~\cite{ren2019generating}, etc. The salience map approach is exemplified by~\cite{zeiler2014visualizing} to test a network with portions of the input occluded to create a map showing which parts of the data have influence on the output. 
For example, Class Activation Mapping (CAM,~\cite{zhou2016learning}) modifies image classification CNN architectures by replacing fully-connected layers with convolutional layers and global average pooling,  thus achieving class-specific feature maps. Grad-CAM~\cite{selvaraju2017grad} generalizes CAM by visualizing the linear combination of the last convolutional layer's feature map activations and label-specific weights, which are calculated by the gradient of prediction w.r.t the feature map activations.



\begin{figure*}[t!]
  \begin{center}
    \includegraphics[width=0.96\textwidth]{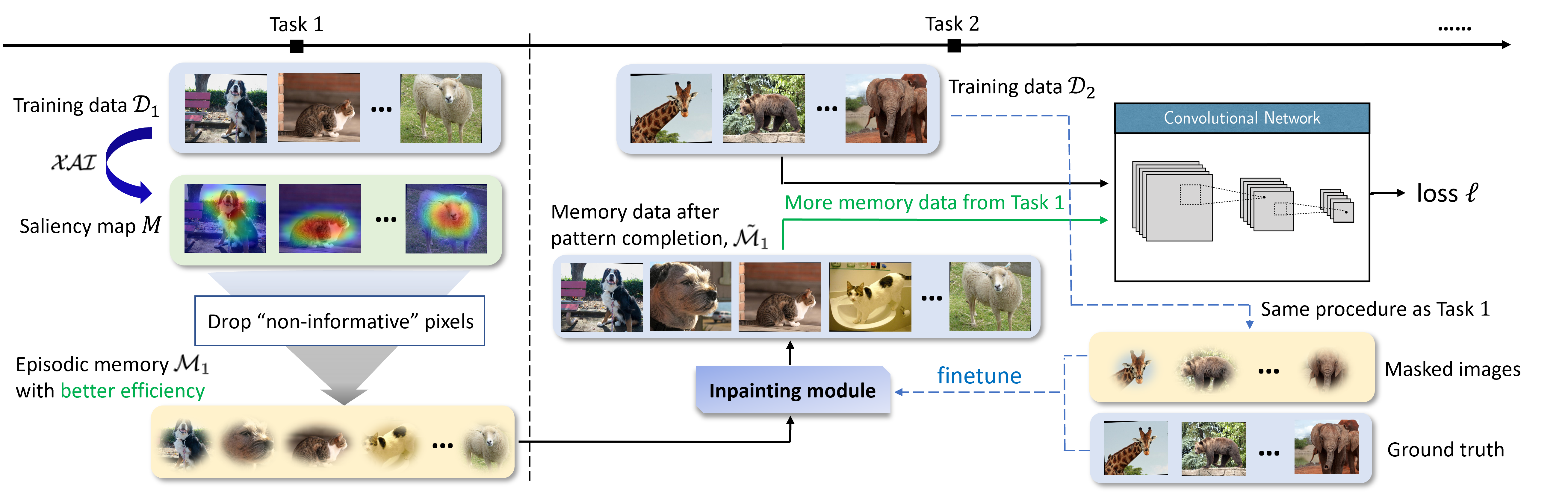}
  \end{center}
  \vspace{-3mm}
  \caption{\textbf{An illustrative overview of the proposed Saliency-Augmented Memory Completion (SAMC) architecture with two tasks.} At task 1, we generate saliency map to extract the informative regions in each image and only store them into episodic memory. At task 2, we apply inpainting model to recover masked images and feed them together with current data into the model for training. We follow same procedure as at task 1 for image extraction and utilize both masked images and ground truth to finetune our inpainting model.}
  \label{fig:SRDML architecture}
  \vspace{-3mm}
\end{figure*}

\section{Problem Formulation}

Continual learning is defined as an online supervised learning problem. Following the learning protocol in~\cite{lopez2017gradient}, we consider a training set $\mathcal{D}=\{\mathcal{D}_1,\mathcal{D}_2,\cdots,\mathcal{D}_T\}$ consisting of $T$ tasks, where $\mathcal{D}_t = \{(\mathbf{x}_{i}^{(t)},\mathbf{y}_{i}^{(t)})\}_{i=1}^{n_t}$ contains $n_t$ input-target pairs $(\mathbf{x}_{i}^{(t)},\mathbf{y}_{i}^{(t)})\in\mathcal{X}\times\mathcal{Y}$. While each learning task arrives sequentially, we make the assumption of \textit{locally i.i.d}, i.e., $\forall\; t, (\mathbf{x}_{i}^{(t)},\mathbf{y}_{i}^{(t)})\overset{iid}{\sim}P_{t}$, where $P_t$ denotes the data distribution for task $t$ and $i.i.d$ for \textit{independent and identically distributed}.

Given such a stream of tasks, our \textbf{goal} is to train a learning agent $f_{\theta}: \mathcal{X}\rightarrow\mathcal{Y}$, parameterized by $\theta$, which can be queried \textit{at any time} to predict the target $\mathbf{y}$ given associated unseen input $\mathbf{x}$ and task id $t$. Moreover, we require that such a learning agent can only store a small amount of seen samples in an episodic memory $\mathcal{M}$ with fixed budget. Under the goal, we are interested in how to achieve \textit{continual learning without forgetting} problem setting defined as following: Given predictor $f_{\theta}$, the loss on the episodic memory of task $k$ is defined as  
\begin{equation}
    \ell(f_{\theta},\mathcal{M}_k)\coloneqq\vert\mathcal{M}_k\vert^{-1}\sum_{(\mathbf{x}_i,k,\mathbf{y}_i)}\phi(f_{\theta}(\mathbf{x_i},k),\mathbf{y}_i), 
\end{equation}
$\forall\;k<t$, where $\phi$ can be \textit{e.g.} cross-entropy or MSE. We consider \textit{constrained optimization}~\cite{delange2021continual} to avoid the losses from increasing, which in turn allows the so called \textit{positive backward transfer}~\cite{lopez2017gradient}. More specifically, when observing the triplet $(\mathbf{x},\mathbf{y},t)$ from the current task $t$, we solve the following \textit{inequality-constrained} problem: 
\begin{equation}
\vspace{-1mm}
\begin{split}
&\text{min}_{\theta}\; \ell\big(f_{\theta}(\mathbf{x},t),\mathbf{y}\big), \;\;\; \text{\,s.t.}, \\
  &\ell\big(f_{\theta},\mathcal{M}_k\big)\leq \ell\big(f_{\theta}^{t-1},\mathcal{M}_k\big),
\end{split}
\label{eq:gem obj 1}
\end{equation}
where $f_{\theta}^{t-1}$ denotes the predictor state at the end of learning task $t-1$ and $k=1,2,\cdots,t-1$. After the training of task $t$, a subset of training samples will be stored into the episodic memory, i.e., $\mathcal{M} = \mathcal{M}\cup\{(\mathbf{x}_{i}^{(t)},\mathbf{y}_{i}^{(t)})\}_{i=1}^{m_t}$, where $m_t$ is the memory buffer size for the current task. During the training of task $t+1$, previously stored samples will serve as $\mathcal{M}_t$ in Eq.~\ref{eq:gem obj 1}.

Our goal above poses significant challenges to existing work: 1). The performance of replay-based methods highly depends on the memory size. Existing work typically considered storing entire samples into the memory, which was inefficient in practice. 2). Memorizing all the details could be problematic, and how to capture the most important and generalizable knowledge through episodic memory is under-explored. 3). Interpretability and transparency over $f_{\theta}$ are typically ignored in existing work, which results in an obscure model with insufficient explanations on how the model is reasoning towards its predictions and how knowledge is stored and refined through the entire continual learning process.

\section{Proposed Method}

In this section, we introduce our proposed Saliency-augmented Memory Completion (SAMC) that addresses all the challenges mentioned earlier and thus narrows the gap between AI and human learning. We innovatively utilize Explainable AI ($\mathcal{XAI}$) methods to select the most salient regions for each image and only store the selected pixels in our episodic memory, thus achieving controllable and better memory efficiency. During the training phase, we leverage learning-based image inpainting techniques for memory completion, where each partially stored image will be restored by the inpainting model. A bilevel optimization algorithm is designed and allows our entire framework to be trained in an end-to-end manner. Our theoretical analyses show that, as long as we properly select the salient region for each image and the inpainting model is well trained, we can simultaneously achieve better memory efficiency and positive backward transfer with a theoretical guarantee.

\subsection{Saliency-augmented Episodic Memory.}

In this section, we demonstrate how to leverage the saliency map generated by saliency methods to select the ``informative'' region of each image and how we design the episodic memory structure to store the images efficiently.

Saliency-based methods generate visual explanation via saliency map, which can be regarded as the impact of specific feature map activations on a given prediction. Given an input image $I\in\mathbb{R}^{H\times W\times C}$, a classification ConvNet $f_{\theta}$ predicts $I$ belongs to class $c$ and produces the class score $f_{\theta}^{c}(I)$ (short as $f_{\theta}^{c}$). The saliency map $M_{I}\in\mathbb{R}^{H\times W}$ is generated by assigning high intensity values to the relevant image regions that contribute more to the model's prediction, i.e.,
\vspace{-1mm}
\begin{equation}
    M_{I} = \Phi(I,f_{\theta},c),
\label{eq:gradcam general}
\vspace{-1mm}
\end{equation}
where $\Phi$ can be any visual explainer that can generate saliency map as Eq.~\ref{eq:gradcam general}. In this paper, we consider Grad-CAM as our choice of $\Phi$, due to that it can pass important "sanity checks" regarding their sensitivity to data and model parameters~\cite{adebayo2018sanity}, which differentiate Grad-CAM~\footnote{More details on Grad-CAM can be found in the appendix.} with many other explanation methods~\cite{ebrahimi2020remembering}.

Suppose we are training on task $t$ with predictor state $f_{\theta}^{t-1}$. Given a mini-batch $\mathcal{B}_t = \{(\mathbf{x}_{i}^{(t)},\mathbf{y}_{i}^{(t)})\}_{i=1}^{bsz}$ from the current task's data $\mathcal{D}_t$ ($bsz$ denotes mini-batch size), we generate the saliency map $M$ following Eq.~\ref{eq:gradcam general}. Intuitively speaking, pixels with higher magnitude in saliency map are more likely to be involved in the region of the target class object while those with lower magnitude are more likely to be the non-target objects or background regions. Specifically, for any input pair $(x,y)\in \mathcal{B}_t$, the masked image $x^{\prime}$ is
\begin{equation}\label{eq:pixel extration}
    x^{\prime}= \mathbbm{1}\{ M_{x}>\mu \} = \mathbbm{1}\{ \Phi(x,f_{\theta},y)>\mu \},
\end{equation}
where $\mathbbm{1}\{\cdot\}$ is the indicator function and $\mu$ is the threshold for controlling each image's drop ratio. 

We denote the mini-batch after extraction as $\mathcal{B}^{\prime}_{t}$. Due to the sparsity in $x^{\prime}$, it is inefficient to store the entire images in episodic memory. We propose to store extracted images in the format of sparse matrix. Specifically, for each image in $\mathcal{B}^{\prime}_{t}$, we first convert it into Coordinate Format (COO)~\cite{bell2008efficient}, i.e., $[(\text{row}_{x^{\prime}},\text{col}_{x^{\prime}},\text{value}_{x^{\prime}})] = \text{COO}(x^{\prime})$, where the left-hand-side is a list of (rwo,col,value) tuples.
After transforming the images into COO format, we store them into the episodic memory as $\mathcal{M}_t = \mathcal{M}_t \cup \text{COO}(\mathcal{B}^{\prime}_{t})$. We consider a simple way for adding samples into memory similar to~\cite{lopez2017gradient}, where the samples are added into memory following \emph{first-in-first-out} (FIFO) and the last group of samples are stored in memory in the end.

\subsection{Memory Completion with Inpainting.}


As shown in Eq.~\ref{eq:gem obj 1}, when we are training on task $t$ we need to calculate the loss on previous samples stored in episodic memory. However, due to the missing pixels, classification models (e.g., CNN) cannot be operated on such ragged images. There are various existing works on how to retrieve images with missing pixels, and we consider the following two approaches:

\noindent\textbf{Rule-based Inpainting.} First approach is to handle the missing pixels by prescribed rules~\cite{bertalmio2001navier,telea2004image}. One advantage of such methods is that they introduce neither extra model parameters nor training costs. Also, one does not need to worry about any potential forgetting of the inpainting method itself during continual learning since it is purely based on hand-crafted rules. One naive candidate of rule-based inpainting is simply padding all missing pixels with zeros. We have included it as a baseline in this paper for comparison.

\noindent\textbf{Learning-based Inpainting.} An alternative is to utilize learning-based inpainting models, and several deep learning models have demonstrated their state-of-the-art performance on various image inpainting tasks~\cite{yeh2017semantic, peng2021generating}. However, the trade-off of their excellent performance is the potential higher memory cost and computational inefficiency. Moreover, it is also challenging to ensure the model generalization for each task during training without severe forgetting.


In this work, we consider autoencoder~\cite{bengio2013representation} as our learning-based inpainting model since 1) autoencoder has been utilized as the backbone in many state-of-the-art deep learning-based inpainting models~\cite{yeh2017semantic, peng2021generating}, which has proven its capability of learning any useful salient features from the masked image, 2) autoencoder is rather a light-weight model which does not bring heavy burden to our framework. Specifically, the autoencoder $\mathcal{A}_{\xi}(x^{\prime})$ can be divided into two parts: an encoder $p_{\xi_{e}}(z|x^{\prime})$ that is parameterized by $\xi_{e}$, and a decoder $p_{\xi_{d}}(\tilde{x}|z)$ that is parameterized by $\xi_{d}$, where $\xi = \{\xi_e,\xi_d\}$, $x^{\prime}$ is the masked image, $\tilde{x}$ is the recovered image, and $z$ is a compressed bottleneck (latent variable). Inpainting autoencoder aims at characterizing the conditional probability
\vspace{-0.1cm}
\begin{equation}
    \mathcal{A}_{\xi}(x^{\prime}) =  p_{\xi_{d}}(\tilde{x}|z)\cdot p_{\xi_{e}}(z|x^{\prime})
    \label{eq:autoencoder}
\end{equation}
to reconstruct the image $\tilde{x}$ from its masked version $x^{\prime}$. Moreover, $\mathcal{A}_{\xi}(\cdot)$  can be trained with the objective of minimizing the reconstruction loss $\argmin_{\xi}||x^* - \mathcal{A}_{\xi}(x^{\prime})||_2^2$, where $x^*$ is the ground truth image.

Furthermore, to maximize the number of samples into the memory $\mathcal{M}_k$, it is imperative to adopt a relatively large drop ratio $\mu$. However, the generalization power of the model would be affected since the details of the original image can hardly be recovered in $\Tilde x$ under a high drop rate. To ensure the model retains its generalization power under such circumstance, we leverage a novel image inpatinting framework that incorporates both rule-based inpainting $\mathcal{R(\cdot)}$ and the autoencoder $\mathcal{A}_{\xi}(x^{\prime})$ to recover the image from coarse to fine. Specifically, given masked image $x^{\prime}$ as defined in  Eq.~\ref{eq:pixel extration}, we propagate it through rule-based inpainting module $\mathcal{R(\cdot)}$ to get a coarse prediction and then feed the output after rule-based inpainting into autoencoder $\mathcal{A}_{\xi}(\cdot)$ for further refinement, i.e., $\tilde{x} = \mathcal{A}_{\xi}(\mathcal{R}(x^{\prime}))$. If we are training on task $t$, given masked images stored in episodic memory $\mathcal{M}_k, \forall\;k<t$, we convert them from COO format back to standard sparse image tensor and generate the memory with pattern completion $\tilde{\mathcal{M}_t}$ as 
\vspace{-1mm}
\begin{equation}
\vspace{-1mm}
    \tilde{\mathcal{M}_k} = \mathcal{A}_{\xi}\Big(\mathcal{R}\big(\text{COO}^{-1}(\mathcal{M}_k)\big)\Big), \quad\forall\;k<t,
\label{eq:generate M tilde}
\end{equation}
where $\text{COO}^{-1}$ denotes the decoding process from COO format back to sparse matrix.


\subsection{Bilevel Optimization.}

In each incremental phase, we optimize two groups of learnable parameters: 1) the model parameter of continual learning classifier, $\theta$, and 2) the inpainting model's parameter $\xi$. Directly optimizing the entire framework could be hard since $\theta$ and $\xi$ are coupled. In this paper, we consider formulating the problem as bilevel optimization, i.e., 
\begin{equation}
\vspace{-1mm}
\begin{split}
    &\text{Upper:\quad} \xi^* = \text{argmin}_{\xi} \lVert \mathbf{x}-\mathcal{A}_{\xi}(\mathcal{R}(\mathbf{x}^{\prime}))\rVert_{2}^{2} \\
    &\text{Lower:\quad}\text{s.t.}\quad \theta^* = \text{argmin}_{\theta}\; \ell\big(f_{\theta}(\mathbf{x},t),\mathbf{y}\big) \\
    & \quad\quad\quad\quad\quad\quad\text{s.t.}\; \ell(f_{\theta},\tilde{\mathcal{M}_k})\leq \ell(f_{\theta}^{t-1},\tilde{\mathcal{M}_k}),
\end{split}
\vspace{-1mm}
\label{eq:bilevel}
\end{equation}
where $k<t$ and $\mathbf{x}^{\prime} = \mathbbm{1}\{ \Phi(\mathbf{x},f_{\theta^*},\mathbf{y})>\mu \}$. The upper-level corresponds to minimizing the image reconstruction loss on the current task, where each raw image is masked by saliency map and then inpainted by the inpainting model. The lower-level corresponds to the standard objective as Eq.~\ref{eq:gem obj 1} for training continual learning classifier $f_{\theta}$ while the constraints are based on inpainted images $\tilde{\mathcal{M}}_k$ instead of actual images $\mathcal{M}_k$.

\noindent\textbf{Optimizing $\theta$:} Given $\xi$, the objective function over $\theta$ becomes a constrained optimization problem. To handle the inequality constraints, we follow~\cite{lopez2017gradient} but with a clearer theoretical explanation.
\vspace{-0.1cm}
\begin{Definition}
Given loss function $\ell$ and an input triplet $(x,y,t)$ or episodic memory after completion $\tilde{\mathcal{M}}_k$, define the loss gradient vector as
\vspace{-0.1cm}
\begin{equation}
    g \coloneqq \partial\ell(f_{\theta}(\mathbf{x},t),\mathbf{y})/\partial\theta,\quad \tilde{g}_k \coloneqq \partial\ell(f_{\theta},\tilde{\mathcal{M}}_k)/\partial\theta .
\label{eq:loss gradient def}
\vspace{-1mm}
\end{equation}
\end{Definition}
The first lemma below proves the \emph{sufficiency} of enforcing the constraint over the loss gradient in order to guarantee \textit{positive backward transfer} as in Eq.~\ref{eq:gem obj 1}.

\begin{lemma}
\label{lem:grad and constraint}
$\forall\; k < t$, with small enough step size $\alpha$, 
\begin{equation}
    \langle g,\tilde{g}_k \rangle \geq 0 \; \Rightarrow\; \ell(f_{\theta},\tilde{\mathcal{M}}_k)\leq \ell(f_{\theta}^{t-1}),\tilde{\mathcal{M}}_k).
\end{equation}
\end{lemma}
\vspace{-1mm}
Guaranteed by the above lemma, we approximate the inequality constraints in Eq.~\ref{eq:bilevel} by computing the angle between $g$ and $\tilde{g}_k$, i.e., $\langle g,\tilde{g}_k \rangle \geq 0$, $\forall\;k<t$. Whenever the angle is greater than $90^{\circ}$, the proposed gradient $g$ will be projected to the closest gradient $\tilde{g}$ in $\ell_2$-norm that keeps the angle within $90^{\circ}$. For more details please refer to~\cite{lopez2017gradient} due to limited space.

\noindent\textbf{Optimizing $\xi$:} Following the training procedure of bilevel optimization~\cite{finn2017model}, we take a couple of gradient descent steps over the inpainting model parameter $\xi$ when samples from new task arrive, i.e., 
\vspace{-0.1cm}
\begin{equation}
    \xi \leftarrow \xi -\beta\cdot\nabla_{\xi} \lVert \mathbf{x}-\mathcal{A}_{\xi}(\mathcal{R}(\mathbf{x}^{\prime}))\rVert_{2}^{2},
\label{eq:xi update}
\end{equation}
where $\beta$ is the step size for $\xi$. It has been shown that \emph{early stopping} in SGD can be regarded as implicit regularization~\cite{rajeswaran2019meta}, and we adopt this technique over $\xi$ to help mitigate the forgetting of the inpainting model. We only take a few steps of SGD over $\xi$ that helps control any detrimental semantic drift of the autoencoder and thus alleviate potential forgetting.

The next lemma shows that the difference between continual learning model's outputs over the original image and inpainted one can be upper bounded by factors related to Grad-CAM's saliency map and the inpainting model's reconstruction error.

\begin{lemma}
\label{lem:err bound over output}
Given input $x\in\mathbb{R}^d$, suppose $\tilde{x}$ is the inpainted sample generated by our proposed method. Denote $y_{x}^{c}$ as $f_{\theta}$'s prediction over $x$ that belongs to class $c$, and $\Delta x\coloneqq \max_{1\leq i\leq d} \vert x_i - \tilde{x}_i \vert$, then: 
\begin{equation}
    err(x,\tilde{x})\coloneqq \vert y_{x}^{c} - y_{\tilde{x}}^{c}\vert \leq \mu\cdot\Delta x\cdot d^{-2}.
\label{eq:error over output}
\vspace{-0.1cm}
\end{equation}
\end{lemma}

Our main theorem below guarantees that, with proper choice of the threshold $\mu$ for pixel dropping and inpainting model with bounded reconstruction error, enforcing the constraint over the memory after pattern completion $\tilde{\mathcal{M}}$ is equivalent to that over the memory with ground-truth images.

\begin{theorem}
\label{thm:1}
Suppose $\ell$ is smooth with Lipschitz constant $L$, and the first order derivative of $\ell$, i.e., $\partial\ell / \partial\theta$ is Lipschitz continuous with constant $L_{\theta}$. We have:
\vspace{-0.1cm}
\begin{equation}
    \langle g,\tilde{g}_k \rangle > 0 \; \Rightarrow\; \langle g,g_k \rangle > 0,\quad as\;\; \tilde{x}\rightarrow x,
\label{eq:grad sufficiency}
\vspace{-0.1cm}
\end{equation}
where $g_k$ is the gradient vector on ground-truth images.
\end{theorem}
\vspace{-2mm}
All formal proofs can be found in the appendix.

\begin{table*}[t]
\centering
\small
\caption{Performance comparison (\%) on image classification datasets. The means and standard deviations are computed over five random runs. When used, episodic memories contain 10 samples per task.}
\vspace{-1mm}
    \begin{tabular}{lcccccc}
    \hline
    \multirow{2}{*}{\textbf{Model}} & \multicolumn{2}{c}{\textbf{Split CIFAR-10}} & \multicolumn{2}{c}{\textbf{Split CIFAR-100}} & \multicolumn{2}{c}{\textbf{Split mini-ImageNet}}\\  \cline{2-7}
           & ACC $(\uparrow)$    &   BWT  $(\uparrow)$   &  ACC $(\uparrow)$     &  BWT $(\uparrow)$  &  ACC $(\uparrow)$   &  BWT  $(\uparrow)$  \\ 
    \hline
    finetune     & 63.24 $\pm$ 2.03  &   -17.91 $\pm$ 1.03       &  41.06 $\pm$ 3.00   &   -17.36 $\pm$ 2.54     & 33.31 $\pm$ 1.67     &  -16.40 $\pm$ 0.32  \\
    \hline
    EWC   & 65.64 $\pm$ 1.79  &   -18.37 $\pm$ 0.98       &  42.22 $\pm$ 1.69   &   -16.54 $\pm$ 2.94      &32.66 $\pm$ 1.91  &  -14.83 $\pm$ 0.67   \\
    LwF & 67.22 $\pm$ 1.68  &   -16.78 $\pm$ 0.73       &  43.79 $\pm$ 1.53   &   -16.23 $\pm$ 2.16  & 34.17 $\pm$ 1.85  &  -13.72 $\pm$ 0.55   \\
    \hline
    iCaRL & 72.53 $\pm$ 1.42  &   -12.43 $\pm$ 0.79       &  48.78 $\pm$ 1.01   &   -15.94 $\pm$ 1.18  & 40.03 $\pm$ 1.62  &  -12.16 $\pm$ 0.42   \\
    GEM    & 70.82 $\pm$ 1.55  &   -15.82 $\pm$ 0.69       &  50.28 $\pm$ 1.28   &   -13.56 $\pm$ 1.11      & 40.78 $\pm$ 1.98  &  -9.87 $\pm$ 0.36 \\
    ER    & 72.13 $\pm$ 1.78 & -12.78 $\pm$ 0.83 & 49.72 $\pm$ 1.82 & -13.84 $\pm$ 2.59 & 40.04 $\pm$ 1.89 & -13.43 $\pm$ 0.53    \\
    MER    & 71.09 $\pm$ 1.32 & -13.69 $\pm$ 0.56 & 47.37 $\pm$ 2.31 & -15.23 $\pm$ 1.27 & 39.63 $\pm$ 1.77 & -11.93 $\pm$ 0.48   \\
    \hline
    EBM    & 72.20 $\pm$ 1.77 & -12.26 $\pm$ 0.61 & 51.29 $\pm$ 1.28 & -11.03 $\pm$ 1.29 & 39.52 $\pm$ 1.48 & -11.75 $\pm$ 0.44   \\
    EEC    & 73.48 $\pm$ 1.91 & -12.13 $\pm$ 0.57 & 53.03 $\pm$ 1.65 & -10.69 $\pm$ 1.32 & 37.32 $\pm$ 1.99 & -12.46 $\pm$ 0.56   \\
    \hline
    \textbf{SAMC (ours)}    & \textbf{75.37 $\pm$ 1.21} & \textbf{-11.12 $\pm$ 0.63} & \textbf{56.16 $\pm$ 1.01} & \textbf{-8.24 $\pm$ 1.47} & \textbf{46.98 $\pm$ 1.42} & \textbf{-4.12 $\pm$ 0.21}    \\
    \hline
    \end{tabular}%
    \vspace{-1mm}
    \label{tab:results_clf}
\end{table*}

\begin{table*}[t]
\scriptsize
\centering
\caption{Comparison of different pattern completion methods. Episodic memories contain 10 samples per task.}
\vspace{-1mm}
    \begin{tabular}{lcccccc}
    \hline
    \multirow{2}{*}{\textbf{Method}} & \multicolumn{2}{c}{\textbf{Split CIFAR-10}} & \multicolumn{2}{c}{\textbf{Split CIFAR-100}} & \multicolumn{2}{c}{\textbf{Split mini-ImageNet}}\\  \cline{2-7}
           & ACC $(\uparrow)$    &   BWT  $(\uparrow)$   &  ACC $(\uparrow)$     &  BWT $(\uparrow)$  &  ACC $(\uparrow)$   &  BWT  $(\uparrow)$   \\ 
    \hline
    Zero-padding    & 68.43 $\pm$ 1.69  &   -16.73 $\pm$ 0.83       &  47.91 $\pm$ 1.41   &   -14.32 $\pm$ 0.93      & 38.73 $\pm$ 1.79  &  -11.62 $\pm$ 0.53 \\
    Rule-based inpainting    & 73.63 $\pm$ 1.43 & -12.93 $\pm$ 0.45 & 54.27 $\pm$ 1.23 & -10.26 $\pm$ 0.85 & 44.10 $\pm$ 1.51 & -6.72 $\pm$ 0.32    \\
    Autoencoder (w./o rule-based)   & 71.02 $\pm$ 1.53 & -13.90 $\pm$ 0.75 & 51.92 $\pm$ 1.68 & -12.73 $\pm$ 1.02 & 42.63 $\pm$ 1.81 & -7.83 $\pm$ 0.59   \\
    Autoencoder (w./ rule-based)    & \textbf{75.37 $\pm$ 1.21} & \textbf{-11.12 $\pm$ 0.63} & \textbf{56.16 $\pm$ 1.01} & \textbf{-8.24 $\pm$ 1.47} & \textbf{46.98 $\pm$ 1.42} & \textbf{-4.12 $\pm$ 0.21}    \\
    \hline
    \end{tabular}%
    \vspace{-0.2cm}
    \label{tab:inpaint comparison}
\vspace{-0.1cm}
\end{table*}

\subsection{Memory Complexity.}

The complexity to store samples into the memory is $\mathcal{O}(T\cdot(1-\bar{\mu})\cdot\lvert\mathcal{M}_t\rvert)$, where $T$ is number of tasks, $\bar{\mu}$ is the average pixel drop ratio determined by $\mu$ in Eq.~\ref{eq:pixel extration}, and $\lvert\mathcal{M}_t\rvert$ is the episodic memory size for each task. The greater the $\bar{\mu}$, the lower the memory complexity to store the same amount of samples. If we consider learning-based inpainting method and denote the number of parameters of the inpainting model as $S$, the overall memory cost becomes $\mathcal{O}(T\cdot(1-\bar{\mu})\cdot\lvert\mathcal{M}_t\rvert + S)$. The cost of storing the inpainting model is often dwarfed by that of storing images, since each image is a high-dimensional tensor and the cost for storing the inpainting model is constant w.r.t $T$.









\section{Experiment}

In this section, we compare our proposed method SAMC with several state-of-the-art CL methods on commonly used benchmarks. More details and additional results can be found in the appendix.~\footnote{Code available at~\url{https://github.com/BaiTheBest/SAMC}}


\subsection{Experiment Setups.} 

\textbf{Datasets.} We perform experiments on three widely-used image classification datasets in continual learning: Split CIFAR-10, Split CIFAR-100, and Split mini-ImageNet~\cite{chaudhry2019tiny}. CIFAR-10 consists of 50,000 RGB training images and 10,000 test images belonging to 10 object classes. Similar to CIFAR-10, CIFAR-100, except it has 100 classes containing 600 images each. ImageNet-50 is a smaller subset of the iLSVRC-2012 dataset containing 50 classes with 1300 training images and 50 validation images per class. For Split CIFAR-10, we consider 5 tasks where each task contains two classes. For Split CIFAR-100 and Split mini-ImageNet, we consider 20 tasks where each task includes five classes. The image size for CIFAR is $32\times 32$ and for mini-ImageNet is $84\times 84$.

\noindent\textbf{Architectures.} Following~\cite{lopez2017gradient}, we use a reduced ResNet18 with three times fewer feature maps across all layers on all three datasets. Similar to~\cite{lopez2017gradient,chaudhry2019tiny}, we train and evaluate our algorithm in `multi-head' setting where a task id is used to select a task-specific classifier head. For learning-based inpainting method, we consider the inpainting autoencoder with three layer encoder and three layer decoder with convolutional structure, respectively. Please refer to appendix for more detail. 


\begin{figure*}[t!]
\centering
  \begin{subfigure}{0.32\textwidth}
    \includegraphics[width=\textwidth]{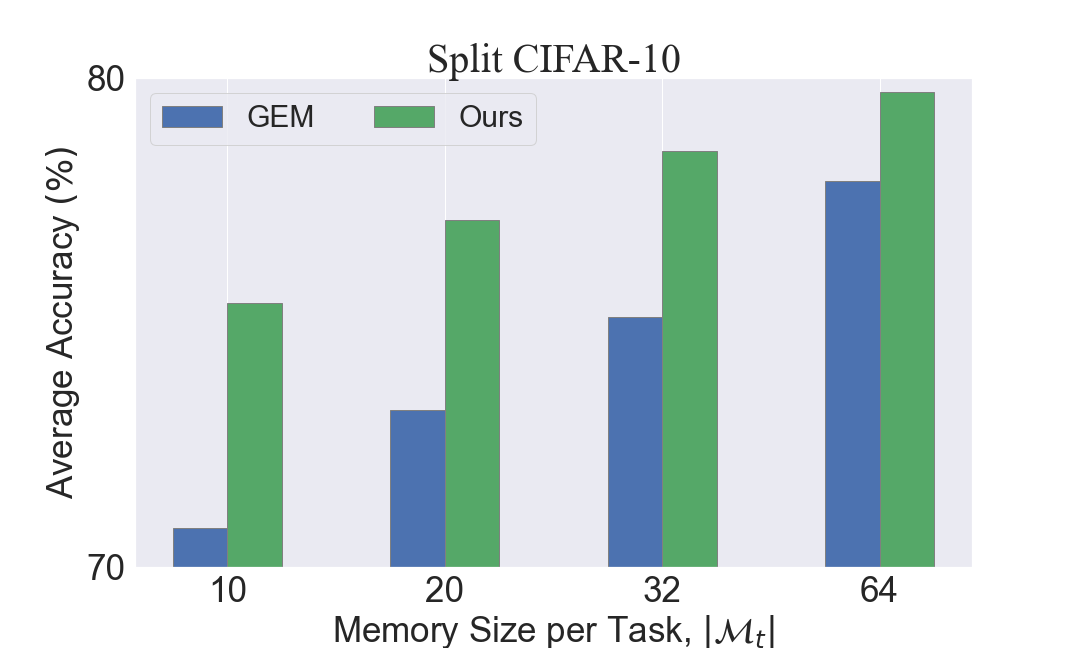}
  \end{subfigure}
  \begin{subfigure}{0.32\textwidth}
    \includegraphics[width=\textwidth]{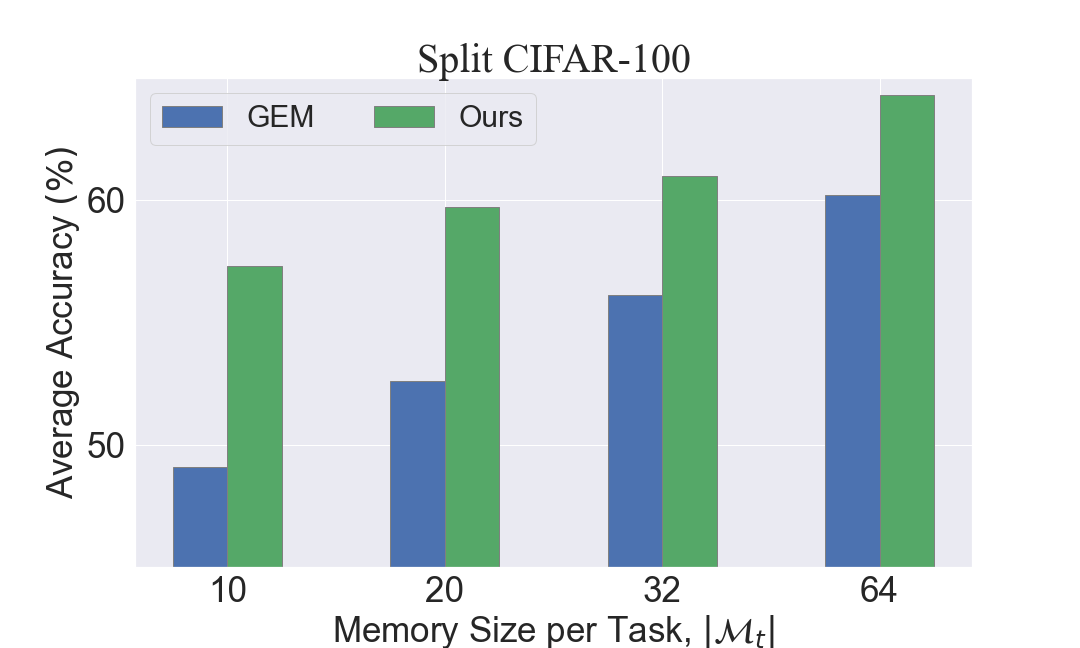}
  \end{subfigure}
  \begin{subfigure}{0.32\textwidth}
    \includegraphics[width=\textwidth]{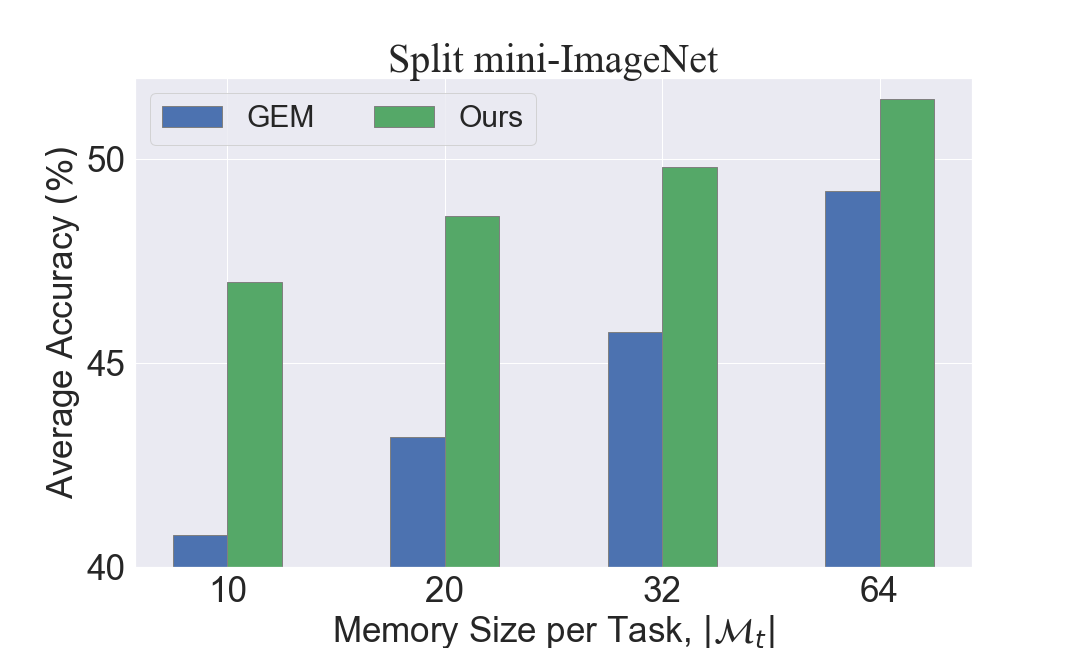}
  \end{subfigure}
  \vspace{-0.2cm}
  \caption{\textbf{Effects of memory size.} We compare ACC for varying memory size per task on Split CIFAR-10, Split CIFAR-100 and Split mini-ImageNet, respectively. Number of tasks for three datasets are 5, 20, and 20.}
  \label{fig:memory size}
\vspace{-0.4cm}
\end{figure*}

\begin{figure*}[t!]
\centering
  \begin{subfigure}{0.23\textwidth}
    \includegraphics[width=\textwidth]{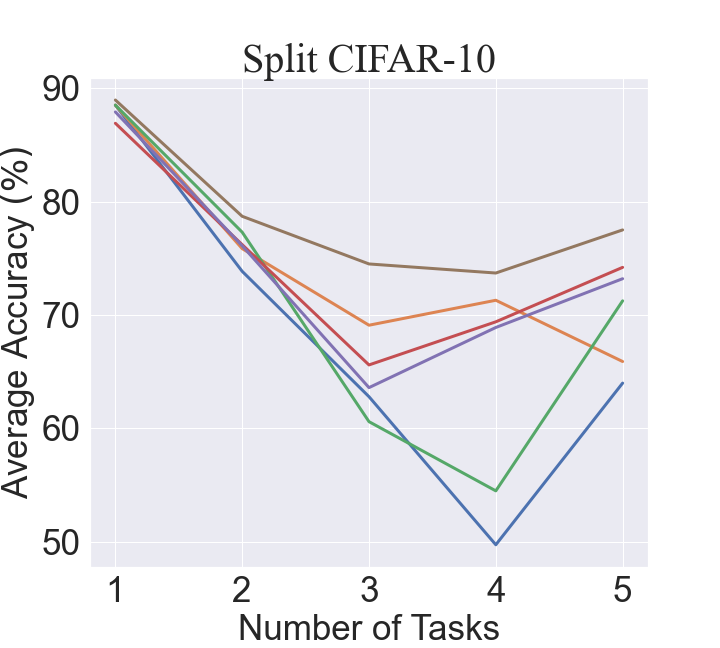}
  \end{subfigure}
  \begin{subfigure}{0.35\textwidth}
    \includegraphics[width=\textwidth]{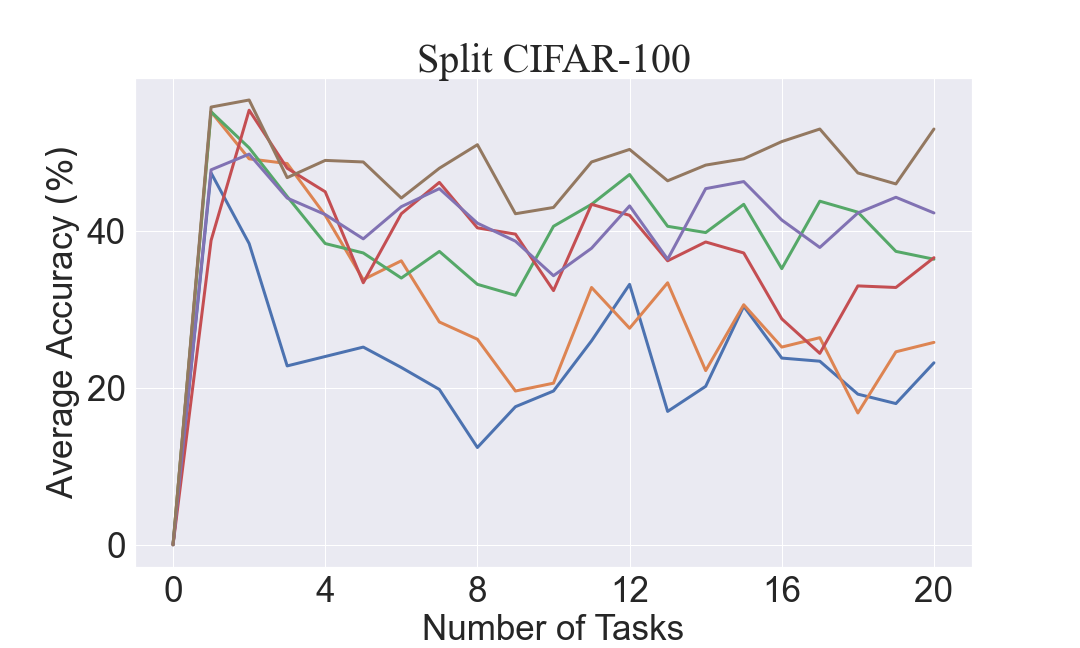}
  \end{subfigure}
  \begin{subfigure}{0.35\textwidth}
    \includegraphics[width=\textwidth]{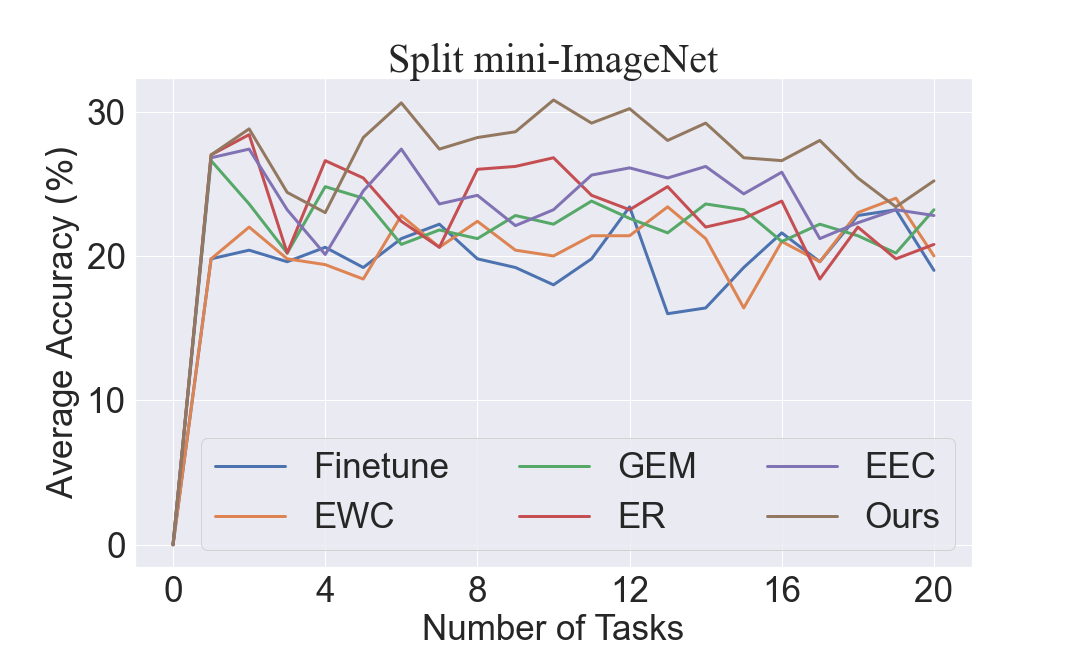}
  \end{subfigure}
  \vspace{-0.2cm}
  \caption{\textbf{Evolution of test accuracy at the first task, as more tasks are learned.} We compare ACC of the first task over the entire training process on Split CIFAR-10, Split CIFAR-100 and Split mini-ImageNet. When used, episodic memories contain 10 samples per task. The legend is same and shown in the right sub-figure only.}
  \label{fig:line chart}
\vspace{-4mm}
\end{figure*}


\noindent\textbf{Comparison Methods.} We compare our method with several groups of continual learning methods, including:
\begin{itemize}[leftmargin=*]
    \item \textbf{Practical Baselines:} \emph{1) Finetune}, a popular baseline, naively trained on the data stream. 
    \item \textbf{Regularization methods:} \emph{2) EWC}~\cite{kirkpatrick2017overcoming}, a well-known regularization-based method; \emph{3) LwF}~\cite{li2017learning}, another regularization-based method for CNNs.
    \item \textbf{Replay-based Methods:} \emph{4) iCaRL}~\cite{rebuffi2017icarl}, a class-incremental learner that classifies using a nearest exemplar algorithm; \emph{5) GEM}~\cite{lopez2017gradient}, a replay approach based on an episodic memory of parameter gradients; \emph{6) ER}~\cite{chaudhry2019tiny}, a simple yet competitive experience method based on reservoir sampling; \emph{7) MER}~\cite{riemer2018learning}, another replay approach inspired by meta-learning. 
    \item \textbf{Pseudo-rehearsal Methods:} \emph{8) EBM}~\cite{li2020energy}, an energy-based method for continual learning; \emph{9) EEC}~\cite{ayub2021eec}, an autoencoder-based generative method.
\end{itemize}

\noindent\textbf{Evaluation Metrics.} We evaluate the classification performance using the ACC metric, which is the average test classification accuracy of all tasks. We report backward transfer, i.e., BWT~\cite{lopez2017gradient} to measure the influence of new learning on the past knowledge. Negative BWT indicates forgetting so the bigger the better. 
We put detailed experimental settings in the appendix.

\subsection{Experiment Results.}

\textbf{Classification Performance.} Table~\ref{tab:results_clf} shows the overall experimental results, where the memory size per task is set to 10 for all datasets, so the total memory buffer size is 50, 200, 200 on CIFAR-10, CIFAR-100 and mini-ImageNet, respectively. On each dataset, our proposed SAMC outperforms the baselines by significant margins, and the gains in performance are especially substantial on more challenging datasets where number of tasks is large and the sample size for each task is small, e.g., CIFAR-100 and mini-ImageNet. Specifically, our model achieves $\sim$12$\%$ and $\sim$15$\%$ relative improvement ratio in accuracy, while achieves $\sim$5$\%$ and $\sim$6$\%$ lower backward transfer compared with the second best method on CIFAR-100 and mini-ImageNet, respectively. This substantial performance improvement can be attributed to better memory efficiency and generalizability provided by SAMC. EWC~\cite{kirkpatrick2017overcoming} performs relatively poor without multiple passes over the dataset and only achieves similar accuracy as finetune baseline. GEM is favored most on CIFAR-100 where it outperformed other methods by some margin. Experience replay methods like ER and MER achieved better performance on CIFAR-100. which is consistent with recent studies on tiny size memory~\cite{chaudhry2019tiny}. Pseudo-rehearsal methods~\cite{li2020energy,ayub2021eec}, though resemble to our proposed method in a way, lack the interpretability and suffer from forgetting in the generative model, so achieve lower accuracy than SAMC.


\begin{figure*}[t!]
\centering
\vspace{-2mm}
  \begin{subfigure}{0.47\textwidth}
    \includegraphics[width=\textwidth]{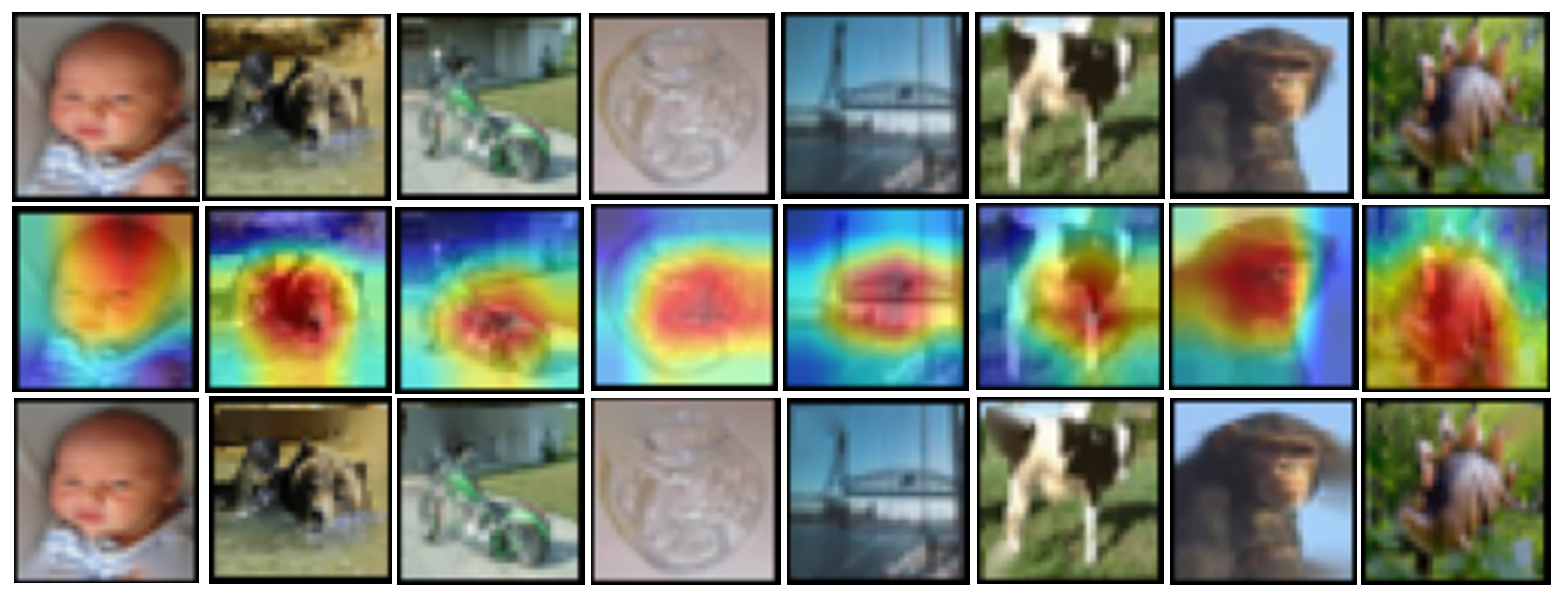}
  \end{subfigure}
  \begin{subfigure}{0.47\textwidth}
    \includegraphics[width=\textwidth]{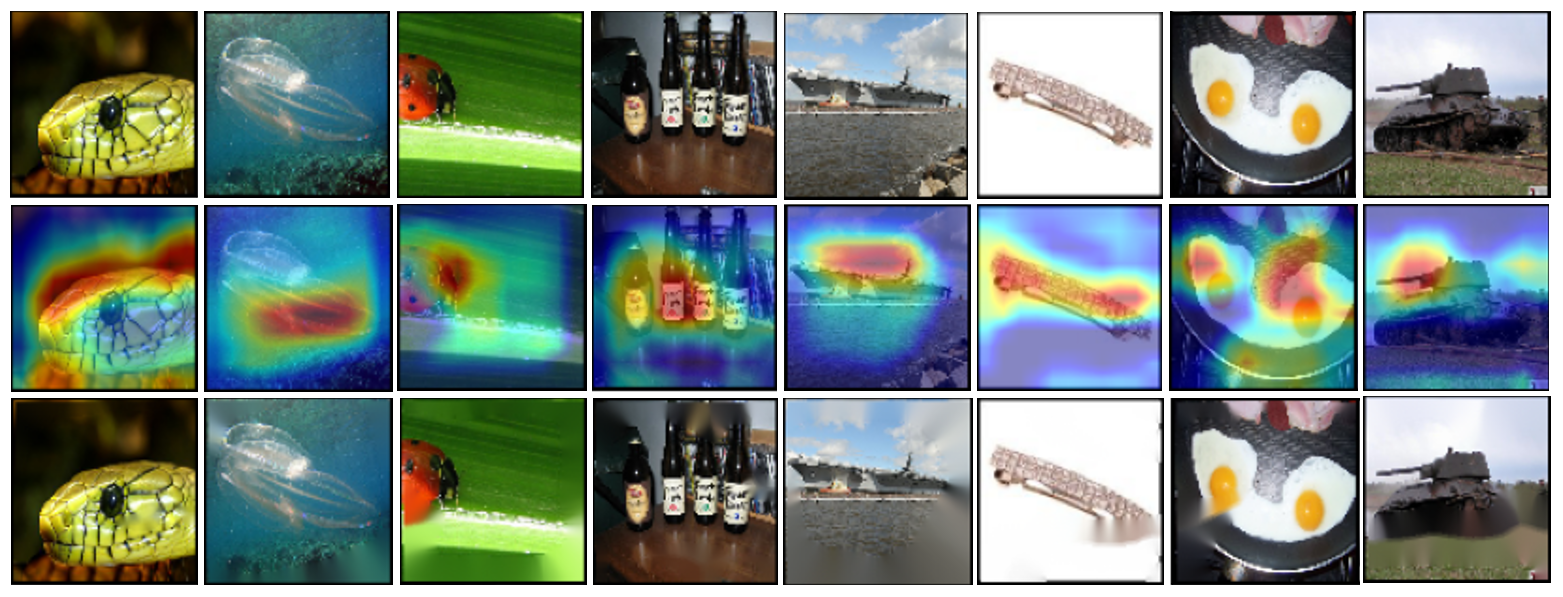}
  \end{subfigure}
  \vspace{-2mm}
  \caption{\textbf{Visualization of saliency map and inpainted images generated by SAMC.} \textbf{Left:} CIFAR-100. \textbf{Right:} mini-ImageNet. \textbf{First row:} Ground truth; \textbf{Second row:} Saliency map; \textbf{Third row:} Inpainted image.}
  \vspace{-3mm}
  \label{fig:quality analysis}
\end{figure*}

\noindent\textbf{Comparison of Pattern Completion Methods.} We investigate the effectiveness of different inpainting techniques on all three datasets. As shown in Table~\ref{tab:inpaint comparison}, naive zero-padding achieves relatively poor performance due to the fact that an image with a large number of pixels with zero value could change the model's prediction dramatically. In fact, our theoretical analysis in Lemma~\ref{lem:err bound over output} tells that the more the inpainted pixel value differs from the original one, the larger $\Delta x$ will be, which results in a more significant prediction error. The rule-based method can achieve consistent results on all the datasets, which may be attributed to its zero forgetting. Autoencoder alone performs a bit worse than rule-based method possibly due to its forgetting, but with the help of rule-based prepcossing and early stopping autoencoder can achieve the best performance.

\noindent\textbf{Effects of Memory Size.} As shown in Figure~\ref{fig:memory size}, we compare our proposed method SAMC with GEM~\cite{lopez2017gradient} over varying memory buffer size. Both GEM~\cite{lopez2017gradient} and SAMC benefit from larger memory size since the accuracy is an increasing function of memory size for both methods. SAMC's improvement in performance over GEM is more substantial in lower data regime, which is possibly due to that sample size is proportional to the memory size. When sample size is small model may easily overfit thus our extra samples can provide maximal benefits, while when sample size is already large enough to learn the model there is no need for extra data. Our model achieves best performance gain in handling few-shot CL which we believe is the most challenging while meaningful case in practice.

\noindent\textbf{Effectiveness in Handling Forgetting.} As shown in Figure~\ref{fig:line chart}, we demonstrate the evolution of first task's test accuracy throughout the entire training process. As can be seen, the proposed method SAMC (red curve) is almost on top of all the curves for the three datasets. Thanks to the memory efficiency of our algorithm and the generalization ability brought by the inpainting model, the proposed method can mitigate catastrophic forgetting to a higher level than other approaches, even with a very small memory buffer.

\begin{table}[h!]
\scriptsize
\centering
\caption{Memory \& computational cost analyses. Our method can achieve 60\% memory saving with only 7.5\% extra computational cost on mini-ImageNet.} 
\begin{tabular}{lcc}
\hline
 &  Disc Consupmtion (MB) & Running Time (sec)  \\
\hline
GEM  & 44 (model) + 413 (images) & 133 \\
SAMC & 44 + 413*$\frac{1}{3}$ (\textcolor{ForestGreen}{$\downarrow$}) + 0.5 (AE,\textcolor{red}{$\uparrow$}) & 133 + 7 (\textcolor{red}{$\uparrow$}) + 3 (\textcolor{red}{$\uparrow$}) \\
\hline
ratio  &  $\sim$60\% (\textcolor{ForestGreen}{$\downarrow$}) & $\sim$7.5\% (\textcolor{red}{$\uparrow$}) \\
\hline
\end{tabular}
\label{tab:memory consumption}
\vspace{-2mm}
\end{table}

\noindent\textbf{Memory Usage Analysis.} As shown in Table~\ref{tab:memory consumption}, we analyze the disc space and training time required by our model on mini-ImageNet dataset and demonstrate that \emph{our computational bottleneck is negligible compared to our memory saving}. The autoencoder (AE) we use is very light-weight and we apply early stopping for its finetuning. By adding quantitative analyses we found our SAMC can achieve \textbf{60\%} memory saving with only \textbf{7.5\%} extra computational cost. Specifically, on mini-ImgNet, SAMC requires 44 MB for ResNet18, 137 MB for images ($\mu$=0.6) and less than 1 MB for AE, while GEM baseline requires 44 MB for ResNet and 413 MB for real images. Meanwhile, by measuring the training time for the first two tasks in single epoch, SAMC takes 7s and 3s for computation induced by Grad-CAM and autoencoder, respectively, in addition to GEM's 130s training time, which leads to a marginal 7.5\% (=(7+3)/130) computational cost. Notice that Grad-CAM itself does NOT introduce any extra parameter.


\noindent\textbf{Qualitative Analysis.} We visualize the saliency map and inpainted images by SAMC in Figure~\ref{fig:quality analysis}. Both saliency map and inpainted images are visually reasonable. Saliency map localization is more accurate on larger images such as mini-ImageNet. The inpainting model generates high-quality images close to the ground truth though there exist occasional and small obscure regions in some cases.


\section{Conclusion}

This paper proposes a new continual learning framework with external memory
based on memory completion/separation theory in neuroscience. By utilizing saliency map, we extract the most informative regions in each image. We store the extracted images in sparse matrix format, achieving better memory efficiency. Inpainting method is used for recovering images from memory with better generalizability. We propose a bilevel optimization algorithm to train the entire framework and we provide theoretical guarantee of the algorithm. Finally, an extensive experimental analysis on commonly-used real-world datasets demonstrates the effectiveness and efficiency of the proposed model.

\bibliography{samc}
\bibliographystyle{plain}

\appendix
\onecolumn

\begin{center}
    {\Large \textbf{Appendix}} 
\end{center}

This is the appendix of paper \emph{Saliency-Augmented Memory Completion for Continual Learning}. In the appendix, we provide more details on experiment settings, our algorithms, additional empirical results as well as the formal proof for each theoretical conclusion in this paper. The experiments in this paper were performed on a 64-bit machine with 4-core Intel Xeon W-2123 @ 3.60GHz, 32GB memory and NVIDIA Quadro RTX 5000. Code available at~\url{https://github.com/BaiTheBest/SAMC}. Note that the code is subject to remodification.

\section{Experiment Setups}

We perform experiments on three commonly used image classification datasets in continual learning domain: Split CIFAR-10, Split CIFAR-100 and Split mini-ImageNet. 

\begin{itemize}
    \item \emph{Split CIFAR-10} is a variant of the CIFAR-10 dataset proposed by Krizhevsky et al originally. This dataset consists of 5 tasks, each with two consecutive classes. We include all training samples (i.e., 50,000) and all test samples (i.e., 10,000) for our model and baselines. Each image has resolution 32 $\times$ 32.
    \item \emph{Split CIFAR-100} is a variant of the CIFAR-100 dataset, similar to CIFAR-10. This dataset consists of 20 tasks, each with five consecutive classes. We include all training samples (i.e., 50,000) and all test samples (i.e., 10,000) for our model and baselines. Each image has resolution 32 $\times$ 32.
    \item \emph{Split mini-ImageNet} is a variant of the mini-ImageNet dataset, which is originally built for few-shot learning tasks. This dataset consists of 20 tasks, each with five consecutive classes. We include all training samples (i.e., 50,000) and all test samples (i.e., 10,000) for all experiments. Each image has resolution 84 $\times$ 84.
\end{itemize}

In this paper, we compare our proposed SAMC with several STOA replay-based methods as well as a regularization-based method and a single baseline. Specifically,

\begin{itemize}
    \item \emph{Finetune}, a model trained continually without any regularization and episodic memory, with parameters of a new task initialized from the parameters of the previous task.
    \item \emph{EWC}, a regularization-based approach that avoids catastrophic forgetting by limiting the learning of parameters critical to the performance of past tasks, as measured by the Fisher information matrix (FIM).
    \item \emph{iCaRL}, a class-incremental learner that classifies using a nearest exemplar algorithm, and prevents catastrophic forgetting by using an episodic memory. 
    \item \emph{GEM}, a model that uses episodic memory as an optimization constraint to avoid catastrophic forgetting. Since GEM and A-GEM have similar performance, we only show the performance of the former in our experiments.
    \item \emph{ER}, a rehearsal-based method that directly use the average of parameter update gradients from the current task's samples and the samples from episodic memory to update the learning agent. Compared with GEM, ER is more computationally efficient and has been shown to work well with tiny memory buffer. However, rehearsal-based methods may easily overfit the samples stored in episodic memory. In this paper, we consider ER with the Reservoir Sampling.
    \item \emph{MER}, another rehearsal-based model that leverages an episodic memory and uses a loss that approximates the dot products of the gradients of current and previous tasks to avoid forgetting. To make the experimental setting more comparable (in terms of SGD updates) to the other methods, we set the number of inner gradient steps to 1 for each outer meta-update with mini-batch size of 10.
\end{itemize}

\section{Model Architectures}

As mentioned, we use a smaller version of ResNet-18 with three times less feature maps across all layers. The number of filters for each block in the reduced ResNet-18 is [20, 40, 80, 160] from shallow layer to deep layer. Also, on top of the convolutional layers is a final linear classifier per task. In this sense, we are considering \emph{multi-head} setting where each head is responsible for a task's prediction and the task descriptor is provided during training and test phase. This is commonly seen on continual learning datasets where the total number of tasks is pretty large and the training samples per task is relatively small, which makes the single head setting quite challenging. We train all the networks and baselines using plain SGD on mini-batches of 10 samples. 

We employ a generally light-weight autoencoder in our proposed SAMC, where the inpainting autoencoder has three layer encoder and three layer decoder with convolutional structure, respectively. Specifically, the number of filters for convolutional layers in encoder is [8, 16, 32] and that for transposed convolutional layers in decoder is [32, 16, 8], respectively. Each convolutional layer in the encoder is followed by a Batch Normalization layer and a Leaky ReLU activation function.

\section{Algorithm For SAMC}

Here we provide the outline of training algorithm of the proposed method.

\begin{algorithm}[h!]
\caption{SAMC training procedure.}
\begin{algorithmic}[1]
\renewcommand{\algorithmicensure}{\textbf{Return:}}
\renewcommand{\algorithmicensure}{\textbf{Require:}}
    \REQUIRE $f_{\theta},\mathcal{A}_{\xi},\mathcal{R},\mathcal{D}^{train},\mu,\alpha,\beta$
    \STATE \text{// $\alpha,\beta$: step size for $\theta$ and $\xi$}
    \STATE $\mathcal{M}_t\gets\{\}$, $\forall\;t=1,2,\cdots,T$
    \FOR {$t=1$ {\bfseries to} $T$}
        \STATE \text{// memory completion}
        \STATE $\tilde{\mathcal{M}_k}=\mathcal{A}_{\xi}(\mathcal{R}(\text{COO}^{-1}(\mathcal{M}_k)))$, $\forall\; k < t$
        \FOR {\texttt{$\mathcal{B}_t \sim \mathcal{D}^{train}_{t}$}}
    \STATE \text{// update $\theta$}
    \STATE $\theta\leftarrow\text{OPT}_{\theta}(\theta,\mathcal{B}_t,\tilde{\mathcal{M}}_{k<t},\alpha)$
    \FOR{\texttt{$(x,y)\in\mathcal{B}_t$}}
    \STATE $x^{\prime}\leftarrow \mathbbm{1}\{ \text{Grad-CAM}(x,f_{\theta},y)>\mu \}$
    \STATE \text{// update episodic memory.}
    \STATE $\mathcal{M}_t\leftarrow\mathcal{M}_t \cup (\text{COO}(x^{\prime}),y)$
    \ENDFOR
    \ENDFOR
    \STATE \text{// finetune autoencoder, i.e, update $\xi$}
    \STATE $\xi \leftarrow \text{OPT}_{\xi}(\xi,\mathcal{M}_t,\mathcal{D}^{train}_{t},\beta)$
    \ENDFOR
    \STATE \textbf{Return:} $f_{\theta}$
\end{algorithmic}
\label{alg:main}
\end{algorithm}

\section{Details on Grad-CAM}

Grad-CAM generates visual explanation via saliency map, which can be regarded as the impact of specific feature map activations on a given prediction. Given an input image $I\in\mathbb{R}^{H\times W\times C}$, a classification ConvNet $f$ predicts $I$ belongs to class $c$ and produces the class score $f_{c}(I)$ (short as $f_c$). Since deeper layers in CNN capture higher-level semantics, taking gradients of a model output w.r.t the feature map activations from one such layer could identify the importance of higher-level semantics for the model prediction. Denote $A\in\mathbb{R}^{H^{\prime}\times W^{\prime}\times C^{\prime}}$ as the feature map activations of the last convolutional layer, where each feature map $A^{k}$ ($k=1,2,\cdots,C^{\prime}$) has height $H^{\prime}$ and width $W^{\prime}$. Grad-CAM first computes the gradient of the score for class c, $f_c$, with respect to feature map activations $A^k$, i.e., $\partial f_c / \partial A^k$, followed by a global-average-pooling over the width and height dimension to obtain the neuron importance weights $\alpha^{c}_k$:
\begin{equation}
    \alpha^{c}_k = \frac{1}{H^{\prime}}\sum_{i=1}^{H^{\prime}}\frac{1}{W^{\prime}}\sum_{j=1}^{W^{\prime}} \frac{\partial f_c}{\partial A^{k}_{ij}}
\label{eq:gradcam alpha}
\end{equation}
Second, Grad-CAM performs a weighted sum over $A^k$ and follows with a ReLU to obtain:
\begin{equation}
    L_{Grad-CAM}^{c} = ReLU(\textstyle\sum_k \alpha^{c}_k A^k).
\label{eq:gradcam weighted sum}
\end{equation}
Finally, saliency map $M(I)\in\mathbb{R}^{H\times W}$ is generated by upsampling $L_{Grad-CAM}^{c}$ to the same resolution as input images,
\begin{equation}
    M(I)=BI_{\uparrow}(L_{Grad-CAM}^{c})
\label{eq:upsampling}
\end{equation}
where $BI_{\uparrow}$ denotes upsampling via bi-linear interpolation.


\section{Hyper-parameter Selection}

Here we report the hyper-parameter grids applied in our experiment. The optimal values for Split CIFAR-10 (cif10), Split CIFAR-100 (cif100) and Split mini-ImageNet (imgnet) are marked accordingly in parenthesis. Once again, memory size per task is set to 10 for all baselines if employed on all datasets, that is equivalent to 5, 2, 2 images \emph{per class} on CIFAR10, CIFAR-100 and mini-ImageNet, respectively.

\begin{itemize}
    \item \textbf{Finetune} \\ \text{$\quad$} learning rate: [ 0.003, 0.01, 0.03, 0.1 (cif10), 0.3, 1.0 (cif100, imgnet) ]
    \item \textbf{EWC} \\ \text{$\quad$} learning rate: [ 0.003, 0.01, 0.03, 0.1 (cif10), 0.3, 1.0 (cif100, imgnet) ]  \\ \text{$\quad$} regularization coefficient: [ 0.1, 1.0 (cif10, cif100, imgnet), 10, 100 ]
    \item \textbf{iCaRL} \\ \text{$\quad$} learning rate: [ 0.003, 0.01, 0.03, 0.1 (cif10), 0.3, 1.0 (cif100, imgnet) ] \\ \text{$\quad$} regularization coefficient: [ 0.1, 1.0 (cif10, cif100, imgnet), 10, 100 ]
    \item \textbf{GEM} \\ \text{$\quad$} learning rate: [ 0.003, 0.01, 0.03, 0.1 (cif10, cif100, imgnet), 0.3, 1.0 ] \\ \text{$\quad$} memory strength: [ 0.0, 0.1, 0.5 (cif10, cif100, imgnet), 1.0 ] 
    \item \textbf{ER} \\ \text{$\quad$} learning rate: [ 0.003, 0.01, 0.03, 0.1 (cif10, cif100, imgnet), 0.3, 1.0 ]
    \item \textbf{MER} \\ \text{$\quad$} learning rate: [ 0.003, 0.01, 0.03, 0.1 (cif10, cif100, imgnet), 0.3, 1.0 ] \\ \text{$\quad$} meta learning rate: [ 0.003, 0.01, 0.03, 0.1 (cif10, cif100, imgnet), 0.3, 1.0 ]
    \item \textbf{SAMC} \\ \text{$\quad$} learning rate: [ 0.003, 0.01, 0.03, 0.1 (cif10, cif100, imgnet), 0.3, 1.0 ] \\ \text{$\quad$} pixel dropping threshold $\mu$: [ 0.1, 0.2, 0.3, 0.4, 0.5, 0.6 (cif10), 0.7 (cif100, imgnet), 0.8, 0.9, 1.0 ]
\end{itemize}

For more details, please refer to our implementation which will be released upon publication.

\section{Sensitivity Analyses}

\begin{figure}[ht!]
\centering
\begin{subfigure}{0.44\textwidth}
  \includegraphics[width=1\linewidth]{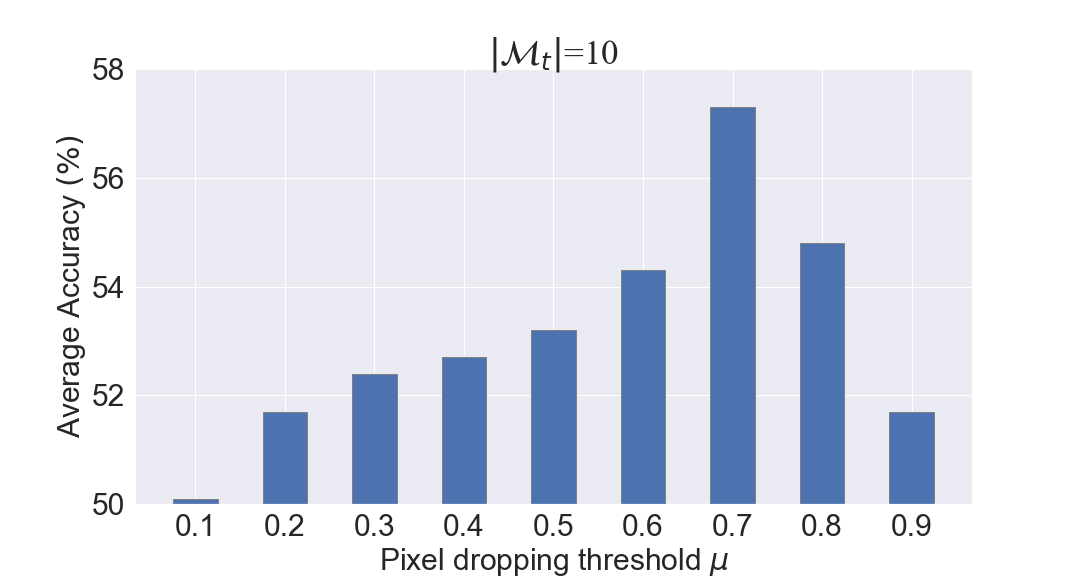}
  \label{fig:Ng1} 
\end{subfigure}
\begin{subfigure}{0.44\textwidth}
  \includegraphics[width=1\linewidth]{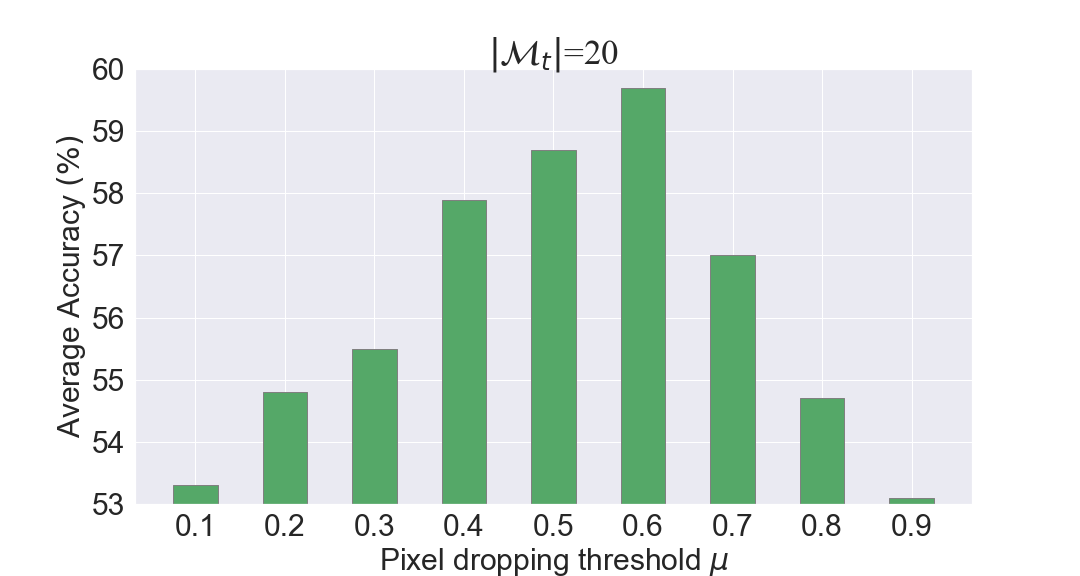}
  \label{fig:Ng2}
\end{subfigure}
\caption{Sensitivity analysis over $\mu$ on Split CIFAR-100.}
  \label{fig:sensitivity analysis}
\end{figure}

We conduct sensitivity analysis on the threshold $\mu$ as in Eq.~\ref{eq:pixel extration}. As can be seen in Figure~\ref{fig:sensitivity analysis}, our method achieves the optimal averaged accuracy when $\mu$ is roughly 0.7 when memory size per task is 10, and the performance decreases as $\mu$ go to either extreme. This pattern is potentially reasonable because as $\mu$ starts increasing from zero, we drop few pixels while memorizing more figures, so the benefits brought by more samples outweigh the error introduced by pixel dropping. However, as $\mu$ becomes larger, the proportion of pixels dropped reaches some level such that the inpainting model cannot reconstruct with fair accuracy, and the error now becomes dominant and outweigh the benefits brought by excessive samples. We also notice that the optimal $\mu$ will change depends on the memory size, where the optimal $\mu$ is around 0.6 when memory size per task increases.

\section{Additional Quantitative Analysis}

\begin{figure*}[h]
\centering
  \begin{subfigure}{0.85\textwidth}
    \includegraphics[width=\textwidth]{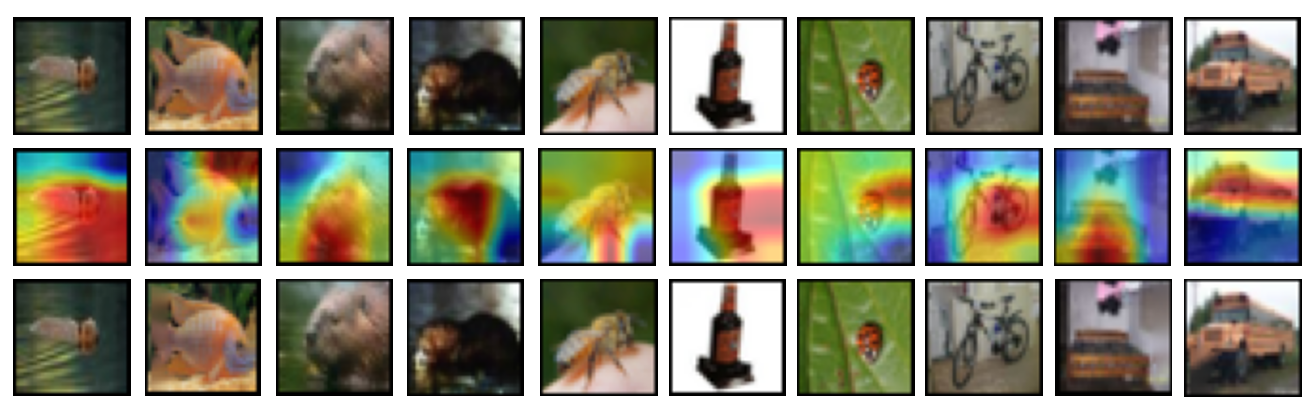}
  \end{subfigure}
  \caption{Visualization of saliency map and inpainted images generated by SAMC on \textbf{Split CIFAR-10.} \textbf{First row:} Ground truth; \textbf{Second row:} Saliency map; \textbf{Third row:} Inpainted image.}
  \label{fig:cifar10 1}
  \vspace{-4mm}
\end{figure*}

In this section, we provide the additional experiment results we put into our appendix due to the page limit of main paper. Refer to Figure~\ref{fig:cifar10 1} for details.

\section{Memory Usage}

Here we report the actual number of images we can store in our episodic memory by leveraging the technique proposed in Section 4.1. Specifically, we report the task average number of images stored in episodic memory on CIFAR-100, with varying $\mu$. The full memory buffer can store 10 raw images per task.

\begin{table*}[h]
\centering
\begin{tabular}{lcccccccccc}
\hline
Pixel dropping threshold $\mu$ & 0 & 0.1 & 0.2 & 0.3 & 0.4 & 0.5 & 0.6 & 0.7 & 0.8 & 0.9 \\
\hline
Task-average images stored & 10 & 11.4 & 12.9 & 15.8 & 18.6 & 22.2 & 30.2 & 43.5 & 71.2 & 112.5 \\
\hline
\end{tabular}
\vspace{-0.1cm}
\caption{Number of images (per task) stored in episodic memory of SAMC on Split CIFAR-100. The default memory size buffer is 10 per task. There are in total 20 tasks in Split CIFAR-100.} 
\label{tab:images stored}
\end{table*}

As can be seen, with larger dropping threshold $\mu$ we can store more samples in the fixed memory buffer. However, combined with our sensitivity analysis in Figure 5, we can find that the optimal performance is achieved in the middle of 0 and 1. The table here provides more insight of our analysis.

\section{Theoretical Proofs}

In this section, we provide the formal proof of theories presented in the paper.

\noindent\textbf{Proof of Lemma 1}

\begin{proof}

Denote $\theta=\theta^{(t-1)}-\alpha\cdot g$, where step size $\alpha>0$, $\theta^{t-1}$ is the state of model parameter of $f_{\theta}$ after task $t-1$.

By Taylor Expansion of $\ell(f_{\theta},\tilde{\mathcal{M}}_k)$ at $\theta^{(t-1)}$,

\begin{equation}
    \begin{split}
        \ell(f_{\theta},\tilde{\mathcal{M}}_k) = \ell(f_{\theta^{(t-1)}},\tilde{\mathcal{M}}_k) + 
        \left\langle \frac{\partial}{\partial\theta}\bigg|_{\theta^{(t-1)}} \ell(f_{\theta},\tilde{\mathcal{M}}_k), \theta - \theta^{(t-1)} \right\rangle + \mathcal{O}(\lVert \theta - \theta^{(t-1)} \rVert^{2})
    \end{split}
\end{equation}
where we expand the loss function up to the first-order, and the last term on right-hand-side is the residual.

By some simple manipulation, we can get:

\begin{equation}
    \begin{split}
        \ell(f_{\theta},\tilde{\mathcal{M}}_k) - \ell(f_{\theta^{(t-1)}},\tilde{\mathcal{M}}_k) &= 
        \left\langle \frac{\partial}{\partial\theta}\bigg|_{\theta^{(t-1)}} \ell(f_{\theta},\tilde{\mathcal{M}}_k), \theta - \theta^{(t-1)} \right\rangle + \mathcal{O}(\lVert \theta - \theta^{(t-1)} \rVert^{2}) \\
        & = -\alpha\cdot \langle g,\tilde{g}_k \rangle + \mathcal{O}(\lVert \theta - \theta^{(t-1)} \rVert^{2}).
    \end{split}
\end{equation}

If we let $\alpha \rightarrow 0^{+}$, which means setting $\alpha$ to approach zero from positive side, the residual will shrink to zero quickly. Hence, we have:
\begin{equation}
    \ell(f_{\theta},\tilde{\mathcal{M}}_k) - \ell(f_{\theta^{(t-1)}},\tilde{\mathcal{M}}_k) \approx -\alpha\cdot \langle g,\tilde{g}_k \rangle.
\end{equation}

\end{proof}

\noindent\textbf{Proof of Lemma 2}

\begin{proof}

Given input $x\in\mathbb{R}^{d}$ and saliency map $M\in\mathbb{R}^d$, we generate $\tilde{x}$ by:
\begin{itemize}
    \item Drop any $x_i$ such that $M_i < \mu$, $i=1,2,\cdots,d$.
    \item Inpaint each element of $x_i$ if it is dropped in step 1.
    \item Copy each element of $x_i$ if it is not dropped in step 1.
    \item Combine step 2 and step 3 gives $\tilde{x}$.
\end{itemize}

Hence, we can find that 
\begin{equation}
    \lVert x - \tilde{x} \rVert^{2}_2 = \sum_{i=1}^d (x_i - \tilde{x}_i)^2 = (\sum_{i\in\text{dropped}} + \sum_{i\in\text{kept}}) (x_i - \tilde{x}_i)^2 = \sum_{i\in\text{dropped}} (x_i - \tilde{x}_i)^2
\label{eq:x and x tilde}
\end{equation}

As a result,
\begin{equation}
    \begin{split}
        err(x,\tilde{x}) &\coloneqq | y_{x}^{c} - y_{\tilde{x}}^{c}  | \\
        &= | f_{\theta}^{c} (x) - f_{\theta}^{c} (\tilde{x}) | \\
        &\approx | \langle \frac{\partial}{\partial\theta}  f_{\theta}^{c} (x), \tilde{x} - x   \rangle | \quad\text{(Taylor Expansion of $f_{\theta}^c$ at $x$)} \\
        &\leq \| \frac{\partial}{\partial\theta}  f_{\theta}^{c} (x)  \| \cdot \| \tilde{x} - x \| \quad\text{(Cauchy-Schwartz Inequality)} \\
        &\leq \mu\cdot \Delta x \cdot \sqrt{d}
    \end{split}
\end{equation}

The last inequality holds since $\frac{\partial}{\partial\theta}  f_{\theta}^{c} (x)$ is essentially the saliency of prediction $f_{\theta}^{c}$ and is bounded by $\mu$ for each dropped position $i$. On the other hand, $\| \tilde{x} - x \|$ is upper bounded by following Eq.~\eqref{eq:x and x tilde} and the definition of $\Delta x$. 

\end{proof}


Now we present the proof of our Theorem 1.

\noindent\textbf{Proof of Theorem 1}

\begin{proof}
First, notice that $\langle g, g_k \rangle = \langle g, g_k + \tilde{g}_k - \tilde{g}_k \rangle = \langle g, g_k - \tilde{g}_k \rangle + \langle g, \tilde{g}_k \rangle$.
Since $\langle g, \tilde{g}_k \rangle > 0$ by our assumption, it suffices to show 
\begin{equation}
    \forall\; \epsilon > 0,\; \text{we can find appropriate}\; \tilde{x},\; \text{s.t.}\; | \langle g, g_k - \tilde{g}_k \rangle | < \epsilon
\end{equation}

By Cauchy-Schwartz Inequality, 
\begin{equation}
    | \langle g, g_k - \tilde{g}_k \rangle | \leq \| g \| \cdot \| g_k - \tilde{g}_k \|
\end{equation}

By Lipschitz smoothness on $\ell$, we have $\| g \| \leq L$. In addition, by Lemma 2 and Lipschitz continuity of $\partial\ell / \partial\theta$, we have $\| g_k - \tilde{g}_k \| \leq L_{\theta}\cdot\mu\cdot\Delta x\cdot\sqrt{d}$. Hence, given $\epsilon>0$, we can always find small enough $\mu$ and $\Delta x$ such that $| \langle g, g_k - \tilde{g}_k \rangle | < \epsilon$.

\end{proof}

\end{document}